\documentclass[11pt]{article}

\usepackage[margin=1in]{geometry}

\usepackage[utf8]{inputenc}
\usepackage[T1]{fontenc}
\usepackage{lmodern}

\usepackage{microtype}

\usepackage{amsmath,amssymb,amsfonts,amsthm}
\usepackage{mathtools}
\usepackage{bm}
\usepackage{makecell}

\usepackage{graphicx}
\usepackage{subcaption}
\usepackage{booktabs}
\usepackage{multirow}
\usepackage{tabularx}
\usepackage{natbib}
\usepackage{adjustbox}

\usepackage[version=4]{mhchem}
\usepackage{siunitx}
\DeclareSIUnit\angstrom{\text{Å}}

\usepackage{enumitem}
\usepackage{hyphenat}

\usepackage{listings}

\usepackage{tikz}

\usepackage[hidelinks]{hyperref}

\usepackage[capitalize,noabbrev]{cleveref}
\crefname{appendix}{Appendix}{Appendices}
\Crefname{appendix}{Appendix}{Appendices}

\usepackage[textsize=tiny]{todonotes}

\theoremstyle{plain}

\theoremstyle{definition}

\theoremstyle{remark}

\newcommand\PACKAGENAME{Landscaper}
\newcommand{\htrace}{\mathrm{tr}(\mathbf{H})}

\usepackage{xcolor}

\usepackage{authblk}

\title{\PACKAGENAME: Understanding Loss Landscapes Through Multi-Dimensional Topological Analysis}

\author[1,*]{Jiaqing Chen}
\author[2,*]{Nicholas Hadler}
\author[1]{Tiankai Xie}
\author[1]{Rostyslav Hnatyshyn}
\author[3]{Caleb Geniesse}
\author[4]{Yaoqing Yang}
\author[2,3,5]{Michael W.~Mahoney}
\author[3]{Talita Perciano}
\author[2,3]{John F.~Hartwig}
\author[1]{Ross~Maciejewski}
\author[3]{Gunther H.~Weber}

\affil[1]{Arizona State University, Tempe, AZ, USA}
\affil[2]{University of California Berkeley, Berkeley, CA, USA}
\affil[3]{Lawrence Berkeley National Laboratory, Berkeley, CA, USA}
\affil[4]{Dartmouth College, Hanover, NH, USA}
\affil[5]{International Computer Science Institute, Berkeley, CA, USA}

\affil[*]{Equal contribution.}

\date{} 

\begin{document}
\maketitle
\begin{abstract}
Loss landscapes are a powerful tool for understanding neural network optimization and generalization, yet traditional low-dimensional analyses often miss complex topological features. 
We present \texttt{\PACKAGENAME}, an open-source Python package for arbitrary-dimensional loss landscape analysis. \texttt{\PACKAGENAME} combines Hessian-based subspace construction with topological data analysis to reveal geometric structures such as basin hierarchy and connectivity. 
A key component is the Saddle-Minimum Average Distance (\textsc{SMAD}) for quantifying landscape smoothness. 
We demonstrate \texttt{\PACKAGENAME}'s effectiveness across various architectures and tasks, including those involving pre-trained language models, showing that \textsc{SMAD} captures training transitions, such as landscape simplification, that conventional metrics miss. 
We also illustrate Landscaper's performance in challenging chemical property prediction tasks, where \textsc{SMAD} can serve as a metric for out-of-distribution generalization, offering valuable insights for model diagnostics and architecture design in data-scarce scientific machine learning scenarios.
\vspace{-6mm}
\end{abstract}

\begin{figure*}[ht]
\centering
\includegraphics[width=\textwidth]{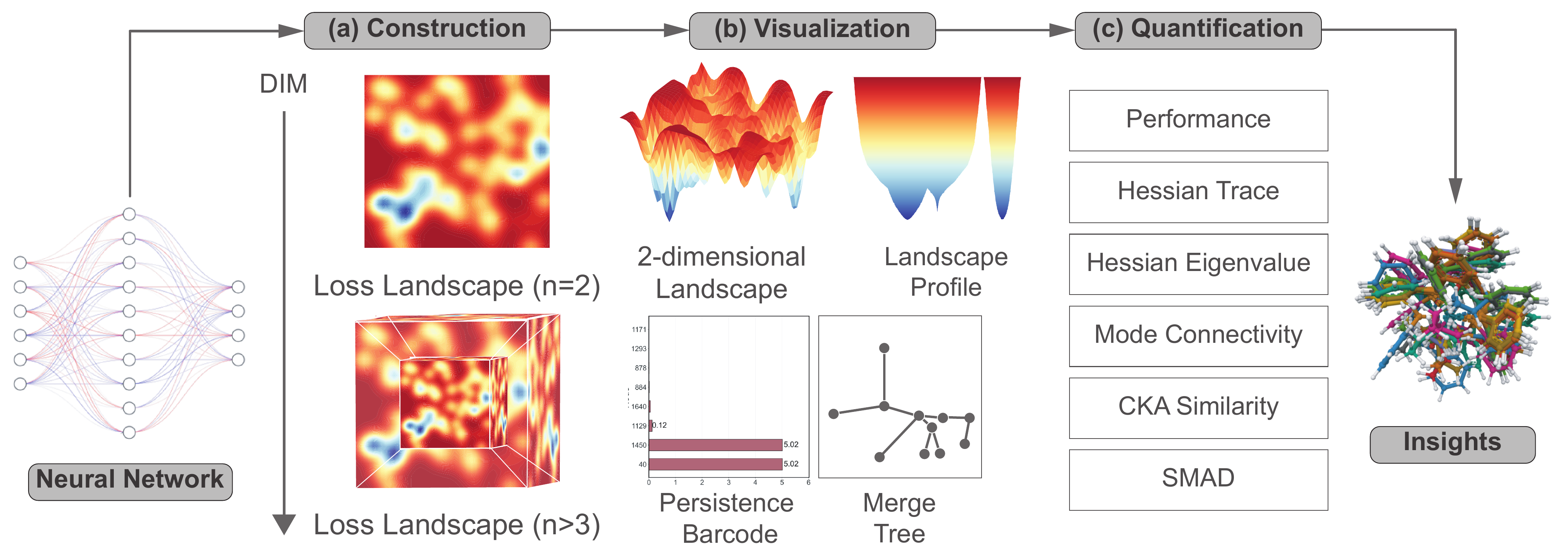}
\caption{
\texttt{\PACKAGENAME}'s workflow involves (a) constructing landscapes in arbitrary-dimensional subspaces; (b) providing visualizations (e.g., landscape profiles, merge trees, and persistence barcodes) to reveal rich geometric/topological features; and (c) quantifying representation diversity and landscape smoothness with metrics (e.g., Hessian trace, \textsc{SMAD}). 
}
\label{fig:method}
\end{figure*}

\section{Introduction}
\label{sec:introduction}

Loss landscape analysis provides insights into the optimization dynamics and generalization performance of machine learning (ML) models that traditional ML metrics alone cannot capture~\citep{goodfellow2014qualitatively,fort2019goldilocks,yang2021taxonomizing}. Of particular interest are metrics designed to tease apart \textit{local structure} from the \textit{global structure} of loss landscapes~\citep{yang2021taxonomizing}. Here, \textit{local structure} refers to the geometry in the immediate neighborhood of a given solution (e.g., curvature, sharpness, and basin shape), whereas \textit{global structure} refers to broader organization across parameter space (e.g., connectivity between basins, barrier heights, and large-scale topology). Common analysis techniques rely on very low-dimensional projections or local curvature metrics (e.g., the Hessian trace)~\citep{li2018visualizing,yao2020pyhessian}. While useful, these approaches can fail to capture the full topological complexity and hierarchical organization of high-dimensional landscapes, often missing interactions between local and global structure that provide critical insight into how models generalize~\citep{cha2021swad,yang2021taxonomizing,iyer2021wideminimadensityhypothesisexploreexploit} and how robust they are to adversarial attack~\citep{kurakin2016adversarial,djolonga2021robustness,yang2022generalized}.

This limitation is particularly acute in scientific machine learning (SciML), where relatively limited data and complex physical constraints often preclude the use of extensive out-of-distribution (OOD) validation sets. In these data-scarce regimes, developing intrinsic geometric metrics, computable solely from training data, can be valuable for assessing model performance and generalization potential.

To address these challenges, we introduce \texttt{\PACKAGENAME}, an open-source Python package designed to bridge the gap between \textit{local} geometry and \textit{global} landscape topology.
\footnote{%
Availble on GitHub, PyPI, and Docker: \url{https://github.com/Vis4SciML/Landscaper}
}
\texttt{\PACKAGENAME} achieves this goal by probing multidimensional loss landscapes via topological data analysis (TDA). 
Crucially, the package implements a diverse analytical suite comprising the full spectrum of metrics required for the landscape taxonomy proposed by~\citet{yang2021taxonomizing}, including Hessian-based curvature, local sharpness, and basin scale estimators~\citep{kwon2021asam, andriushchenko2023modern}. While prior work demonstrates that local and global properties can be inferred from a combination of these traditional metrics~\citep{yang2021taxonomizing}, we introduce a novel, topologically grounded metric to quantify global structure directly: the \emph{Saddle-Minimum Average Distance (\textsc{SMAD})}. 
Unlike the Hessian trace, which strictly measures local curvature, \textsc{SMAD} aggregates the geometric features and overall shape of the loss landscape into a unified scalar that quantifies global smoothness.
\texttt{\PACKAGENAME} also implements novel visualization techniques that project high-dimensional geometric structures into interpretable profiles \citep{geniesse2024visualizing}, providing insights that conventional low-dimensional visualizations fail to capture.

To demonstrate the diagnostic capabilities of \texttt{\PACKAGENAME}, we conduct a comprehensive evaluation across three diverse architectural families: Convolutional Neural Networks (CNNs), Transformers, and Graph Neural Networks (GNNs). 
Specifically, we leverage this framework to investigate optimization geometry in both standard benchmarks and challenging SciML tasks. 
Table~\ref{tab:experiment_summary} outlines the scope of our empirical analysis and the primary insights derived from each case~study.

\begin{table*}[t!]
\centering
\small
\setlength{\tabcolsep}{4pt}
\renewcommand{\arraystretch}{1.25}
\caption{
Summary of empirical evaluations conducted using \texttt{\PACKAGENAME}. The analyses span standard benchmarks and specialized SciML tasks, demonstrating the utility of both \textsc{SMAD} and high-dimensional TDA visualization across different domains.
}
\label{tab:experiment_summary}
\begin{tabularx}{\textwidth}{l l X}
\toprule
\textbf{Model / Architecture} & \textbf{Task / Dataset} & \textbf{Key Insight / Analysis Focus} \\
\midrule
ResNet-20 & CIFAR-10 & Validates \textsc{SMAD} sensitivity, showing that skip connections reduce \textsc{SMAD} by $\approx 40\times$, consistent with theoretical smoothing effects. \\
\addlinespace
CNN & CIFAR-10 & Validates that \textsc{SMAD} correctly captures landscape variation across several training regimes: underfit, well-fit, and overfit. \\
\addlinespace
SchNet; DimeNet++ & QM9 (Molecular Properties) & Hessian trace analysis correlates with performance gaps between isotropic (SchNet) and directional (DimeNet++) architectures. \\
\addlinespace
MultiBERT Suite & Language Pre-training & \textsc{SMAD} decouples local curvature from global topology, capturing landscape simplification missed by Hessian metrics. \\
\addlinespace
DimeNet++ & Chemical Reaction Prediction & (1) High-dimensional TDA resolves connectivity artifacts hidden in 2D projections; (2) \textsc{SMAD} correlates and successfully predicts OOD generalization using only ID landscapes. \\
\bottomrule
\end{tabularx}
\end{table*}

\section{Background}
\label{sec:background}

\paragraph{Loss Landscapes.}
\label{sec:background_landscape}
Loss landscapes provide valuable insights into network architectures and learning dynamics~\citep{goodfellow2014qualitatively,im2016empirical,ballard2017energy,li2018visualizing,yao2020pyhessian,MM19_HTSR_ICML,MM20_SDM,martin2021implicit,martin2021predicting,yang2022evaluating,yang2021taxonomizing,zhou2023three,sakarvadia2024mitigating,khan2024sok}. Some notable examples include demonstrating the robustness of transfer learning~\citep{djolonga2021robustnesstransferabilityconvolutionalneural}, evaluating a model's robustness to adversarial attacks~\citep{zheng_detecting_2023}, and providing deeper insights into a model’s ability to generalize~\citep{cha2021swad}.

A loss landscape can be thought of as a height map, where each point on the grid represents the model being analyzed with perturbed parameters. The value (or height) at each grid point is typically generated by calculating the total or average loss over the dataset for the perturbed model.
Naively perturbing models by modifying parameters individually would yield an extremely high-dimensional landscape, with one dimension per parameter, which is impractical for both analysis and computation. Instead, each axis represents a direction along which all of the model’s parameters are perturbed~\citep{goodfellow2014qualitatively, li2018visualizing}. The top eigenvectors of the Hessian of the loss function with respect to the model’s parameters provide a natural source/choice for these directions~\citep{yao2020pyhessian}. These eigenvectors reflect perturbation directions that affect the loss the most. This approach allows the landscape to span an arbitrary number of dimensions, although typically only the first few eigenvectors are used because the rest are often less visually informative.

\paragraph{Topological Data Analysis.}
\label{sec:tda_background}
A large body of work has demonstrated that the geometry of loss landscapes reflects a model’s performance and its ability to generalize~\citep{xie2024evaluating,xie2025losslens, ly2025optimization, cha2021swad}; however, historically, such analyses have been qualitative, often focusing on comparing loss landscapes visually. To quantify these observations, \citet{xie2024evaluating} proposed the use of topological data analysis (TDA), an applied form of mathematical topology that studies the ``shape" of data. 
Here, we very briefly introduce a few fundamental concepts of persistent homology~\citep{edelsbrunner2008persistent} that are key to understanding these techniques. 

Persistent homology summarizes the global structure of high-dimensional data by tracking which features (groups of points) persist as we vary a scale parameter. 
Given a function $f: V \to \mathbb{R}$, we examine regions of the landscape where the function value is below a threshold $h$. 
While persistent homology can characterize features of various dimensions (e.g., loops or voids), this work focuses exclusively on 0-dimensional persistence, which tracks the number of connected components. 
As $h$ increases, these regions expand outward from local minima (e.g., basins) and eventually merge at saddle points. New topological features appear at minima and merge when they meet at saddle points. 

The significance of each connected component is then quantified by its so-called \emph{persistence} value, defined as the difference between the function values at the minimum (birth) and the saddle point (death). Features with large persistence values correspond to prominent basins separated by high barriers, while short-lived features are typically attributed to noise or small-scale fluctuations. This process can be visualized through merge trees~\citep{contour_tree, heine2016survey}, as well as persistence barcodes (\cref{apd:method_visualization}), which both provide a summary of the landscape's global structure.

\section{Methods}
\label{sec:method}

\texttt{\PACKAGENAME} offers a comprehensive framework for exploring loss landscapes of models of any size through a three-stage workflow: \textit{construction}, \textit{visualization}, and \textit{quantification}.
First, we construct high-dimensional landscapes to capture complex geometry; second, we visualize these structures using merge trees, persistence barcodes, and landscape profiles to gain qualitative insights; and third, we quantify global landscape smoothness using a novel metric, \textsc{SMAD}, alongside complementary measures such as CKA similarity, Hessian trace, and mode connectivity. \cref{apd:method_construction,apd:method_visualization} provide additional implementation details for the construction and visualization stages.

\paragraph{Construction.}
\texttt{\PACKAGENAME} generates loss landscapes by sampling along the top-$n$ Hessian eigenvector directions, which define joint perturbations of \emph{all model parameters} (\cref{sec:background_landscape}). We typically restrict analysis to $2\leq n \leq 5$ directions. This upper bound is imposed by the fact that the number of grid points required to maintain sufficient topological resolution grows exponentially ($k^n$) with dimension, making dense sampling computationally intractable beyond 5D for large models. Higher-dimensional sampling is supported via random or adaptive sampling strategies that avoid exhaustive gridding. 

\paragraph{Visualization.}
\texttt{\PACKAGENAME} bridges TDA with modern ML visualizations to deliver actionable insights. 
Unlike conventional packages that rely solely on scalar metrics, \texttt{\PACKAGENAME} integrates TDA visualizations that allow users to qualitatively explore the loss subspace in detail. In particular, \texttt{\PACKAGENAME} provides the ability to visualize loss landscapes, landscape profiles (\cref{fig:experiments_tda}(a), third and fourth column), merge trees (\cref{fig:experiments_tda}(b), second column), and persistence barcodes (\cref{fig:experiments_tda}(b), third column) to provide varied perspectives on the landscape’s topology.

\paragraph{Quantification.} \texttt{\PACKAGENAME} introduces the saddle–minimum average distance (\textsc{SMAD}), a novel topological metric that quantifies the smoothness of a loss landscape by examining its topological persistence (\cref{sec:tda_background}). \textsc{SMAD} aggregates the landscape's topological persistence following branches of the merge tree (\cref{fig:mergetree_persistencebarcode}) and provides a single interpretable value. 

This quantification is accomplished by first calculating the persistence value for each connected saddle-minimum pair (\cref{sec:background}) and normalizing it by the total range of loss values $R = \max f(x) - \min f(x)$; this normalization facilitates comparisons between landscapes. To safeguard against large basins dominating the final \textsc{SMAD} value, we weigh each persistence value by $w$, the number of points that belong to the minimum's stable manifold~\citep{edelsbrunner2008persistent}. In simple terms, these are the points that lie along the path from the saddle to the minimum. With this in mind, consider a merge tree $T = (V, E)$, a set of nodes $M$ where $v \in M$ \textit{iff} it is a minimum, a weight function $p: E \to \mathbb{R}$, where $p(u,v) = |f(u) - f(v)|$ (persistence), and a set of edges $S \subset E$ consisting of the edges connecting each minimum in $M$ to its parent saddle:
\begin{equation}
  \label{eq:smad}
  \text{SMAD} = \frac{1}{|S|} \sum_{i=1}^{|S|} \left( \frac{p_i}{R} \cdot \frac{w_i}{N} \right) ,
\end{equation}
where \(|S|\) represents the number of valid saddle-minimum persistence pairs (i.e., edges incident to the minima) and $N$ is the total number of points in the loss landscape. 
\textsc{SMAD} provides an intuitive summary of landscape structure: a lower value indicates smoother landscapes with shallow, interconnected basins, while a higher value reveals sharper minima or separated basins. 
The 2D case illustrates how \textsc{SMAD} quantifies basin depth and connectivity, with complex higher-dimensional subspaces exhibiting analogous topological features. 

\section{Results}
\label{sec:experiments}
We examined a number of diverse architectures (\cref{tab:experiment_summary}) to evaluate whether \textsc{SMAD} and other methods in \texttt{\PACKAGENAME} can capture meaningful properties of loss landscapes in ML and SciML. 
These case studies provide strong evidence that \textsc{SMAD} and other metrics in \texttt{\PACKAGENAME} offer insights beyond traditional loss landscape curvature analysis. We accomplished this through three distinct analyses:

\begin{enumerate}[leftmargin=6mm, topsep=0mm, itemsep=0mm]
    \item \textbf{Validation and Cross-Architecture Analysis.} We demonstrate that \texttt{\PACKAGENAME} captures important loss-landscape features and links them to actionable, meaningful ML outcomes across a range of general ML architectures, including CNNs, GNNs, and Transformers, including effects missed by traditional metrics. 
    \item \textbf{The Effectiveness of TDA in Higher-Dimensional Subspaces.} We demonstrate that models with nontrivial loss surfaces cannot be adequately described with traditional low-dimensional visualizations. TDA exposes high-dimensional connectivity, basin hierarchies, and curvature patterns that are essential for understanding model behavior.
    \item \textbf{Assessing ID and OOD Generalization with SMAD.} 
    We show that \textsc{SMAD}, computed solely from the in-distribution (ID) loss landscape, is strongly predictive of and identifies the optimal amount of data augmentation and the best-performing feature engineering strategy for out-of-distribution (OOD) performance in a challenging SciML task, namely a catalyst selectivity prediction task. This behavior provides evidence that \textsc{SMAD} could act as a quantifiable metric for generalization, capable of guiding model development and feature design strategies in ML and SciML.
\end{enumerate}

\subsection{Validation and Cross-Architecture Analysis}

\paragraph{Controlled Validation Using Skip Connections.} We first examined a ResNet trained on CIFAR-10, comparing variants with and without skip connections. 
Prior literature has established that skip connections eliminate singularities and smooth out the loss landscape \citep{li2018visualizing}.
Consistent with this expectation, adding skip connections decreased \textsc{SMAD} by approximately 40 $\times$ on average (\cref{table:resnet}). 
This drastic reduction confirms that \textsc{SMAD} correctly identifies the topological smoothing known to facilitate optimization. 
Unlike simple curvature metrics, which can yield conflicting signals depending on the model architecture or training configuration~\citep{yao2020pyhessian}, \textsc{SMAD} consistently reflects improvements in global landscape structure.

\begin{table}[ht]
    \centering
    \caption{Skip connection analysis for ResNet20 on CIFAR-10 averaged across multiple random seeds. Adding skip connections substantially decreases \textsc{SMAD}, consistent with smoother, more optimizable landscapes.}
    \label{table:resnet}
    \begin{tabular}{lcc}
        \toprule
        Skip & \textsc{SMAD} & Persistence Range \\
        \midrule
        Yes & $(1.5 \pm 0.73)\times 10^{-4}$ & $4.5 \pm 1.22$ \\
        No  & $(5.7 \pm 9.8)\times 10^{-3}$  & $3.4 \pm 3.0$ \\
        \bottomrule
    \end{tabular}
\end{table}

\paragraph{Characterizing Underfitting and Overfitting with SMAD.}
We next analyzed three CNNs trained on CIFAR-10 as a controlled testbed spanning underfit, well-fit, and overfit regimes, enabling direct quantitative comparisons of landscape structure across training conditions. The underfit model was trained for only 5 epochs on a 5,000-sample subset using a large batch size (512), high dropout ($p=0.5$), strong weight decay ($\lambda=5\times10^{-2}$), and no data augmentation, conditions designed to constrain learning capacity. The well-fit model was trained for up to 80 epochs on the full training set with standard data augmentation, moderate regularization (dropout $p=0.15$, weight decay $\lambda=1\times10^{-4}$), and early stopping (patience of 10 epochs) to achieve good generalization. The overfit model was trained for 200 epochs on only 1,000 samples with no regularization (dropout $p=0$, weight decay $\lambda=0$) and no augmentation, encouraging the network to memorize the training~data.

The results demonstrate that \textsc{SMAD} varies in a manner consistent with the expected qualitative geometry of the loss landscape (\cref{tab:fit_regime}). The well-fit regime exhibits the lowest \textsc{SMAD} value, suggesting a smoother, more coherent landscape structure around the solution. The underfit regime shows an intermediate \textsc{SMAD} value, consistent with a model that has not reached a similarly stable solution. The overfit regime displays the highest \textsc{SMAD} value, indicating increased global roughness in the sampled landscape, in line with solutions that generalize poorly. These results support that \textsc{SMAD} is a regime-sensitive descriptor of global loss landscape geometry and captures expected trends; in particular, lower \textsc{SMAD} values are consistent with smooth landscapes that have been associated with reduced overfitting and improved generalization~\citep{iyer2021wideminimadensityhypothesisexploreexploit}. The 2D and 3D contour plots for these three models are provided in \cref{apd:cnn_landscapes}.

\begin{table}[ht]
\caption{Train and test accuracy with corresponding \textsc{SMAD} values across three training regimes for a CNN trained on CIFAR-10.}
\centering
\begin{tabular}{cccc}\toprule
Regime&  Train Accuracy&  Test Accuracy& SMAD\\\midrule
Underfit&  0.548&  0.489& 0.234\\
Well-fit&  0.986&  0.879& 0.130\\
Overfit&  1.000&  0.553& 0.294\\ \bottomrule
\end{tabular}
\label{tab:fit_regime}
\end{table}

\paragraph{Loss-Landscape Analysis for Atomistic SciML: Local Curvature vs Global Structure.} 
We then examined two atomistic SciML models, specifically \textit{SchNet} \citep{schuttSchNetContinuousfilterConvolutional2017a} and \textit{DimeNet++} \citep{gasteigerDirectionalMessagePassing2022a}, trained on the QM9 benchmark \citep{ramakrishnanQuantumChemistryStructures2014} to predict the quantum-mechanical properties of molecules. While both architectures encode atomic environments through message-passing convolutional layers, they differ substantially in how geometric information is represented. \textit{SchNet} models local atomic interactions solely via interatomic distances, yielding smooth but direction-agnostic geometric representations. In contrast, \textit{DimeNet++} incorporates directional information, enabling explicit modeling of interatomic angles and capturing more complex local chemical environments. As a result, \textit{DimeNet++} provides more expressive embeddings of molecular structure.

Analyzing the Hessian trace across all QM9 regression targets (\cref{tab:hessian-trace}) revealed a strong negative correlation (Spearman’s $\rho = -0.77$, $p = 0.009$) between inter-model differences in the Hessian trace and the corresponding performance gap. However, \textsc{SMAD} was not significantly correlated with inter-model performance differences. Together, these results suggest that, for these atomistic models, local curvature of the loss landscape is more predictive of accuracy than the global structure captured by SMAD, demonstrating the complementary nature of using both metrics. In particular, smaller Hessian trace values (i.e., smoother, less-curved optimization landscapes) are associated with higher predictive accuracy, whereas larger values (i.e., sharper or more irregular curvature) correspond to worse performance. Concretely, this behavior suggests a simple workflow benefit: when comparing candidate atomistic models, the Hessian trace can serve as an early, model-agnostic indicator for expected accuracy, helping prioritize which architectures or training runs to keep before running full benchmark sweeps.

\begin{table}[ht]
\centering
\caption{Hessian trace ($\htrace$) and loss (MAE) for SchNet and DimeNet++ trained and evaluated on 10 QM9 regression targets.}
\label{tab:hessian-trace}
\begin{tabular}{@{}lrrrr@{}}
\toprule
{\shortstack{QM9\\Target}} &
{\shortstack{SchNet\\$\htrace$}} &
{\shortstack{DimeNet++\\$\htrace$}} &
{\shortstack{SchNet\\MAE}} &
{\shortstack{DimeNet++\\MAE}} \\
\midrule
$\mu$                  & 281.47  &    7.88   &  0.02  &  0.03 \\
$\alpha$               &  75.84  &   14.90   &  0.12  &  0.04 \\
$\epsilon_{HOMO}$      & -20.02  &  -14.19   & 46.11  & 24.43 \\
$\epsilon_{LUMO}$      & -16.54  &   19.92   & 37.95  & 19.42 \\
$\langle R^2 \rangle$  & 133.76  & -8610.65  &  0.16  &  0.29 \\
ZPVE                  &  -0.66  &  -24.28   &  1.58  &  1.22 \\
$U_0$                 &   6.86  &  243.38   & 11.58  &  6.15 \\
$U$                   &  -4.67  &  -77.69   & 11.56  &  6.20 \\
$H$                   &  -3.56  &   13.00   & 11.64  &  6.49 \\
$G$                   &  10.67  &  -69.50   & 12.52  &  7.41 \\
\bottomrule
\end{tabular}
\end{table}

\paragraph{Capturing Training Dynamics in Pre-trained Language Models.} To further demonstrate \texttt{\PACKAGENAME}'s analytical versatility, we examined the MultiBERTs suite \citep{sellam2021multiberts}. We examined loss landscapes across 25 checkpoints, spanning 5 distinct random seeds (0--4) and 5 training steps (0k, 20k, 40k, 100k, 2000k). For each checkpoint, we computed the loss surface on an $11 \times 11$ grid (distance $= 0.5$) projected onto the top-2 Hessian eigenvectors.

\begin{figure}[!ht]
  \centering

  \begin{subfigure}[t]{0.49\columnwidth}
    \centering
    \includegraphics[width=\columnwidth]{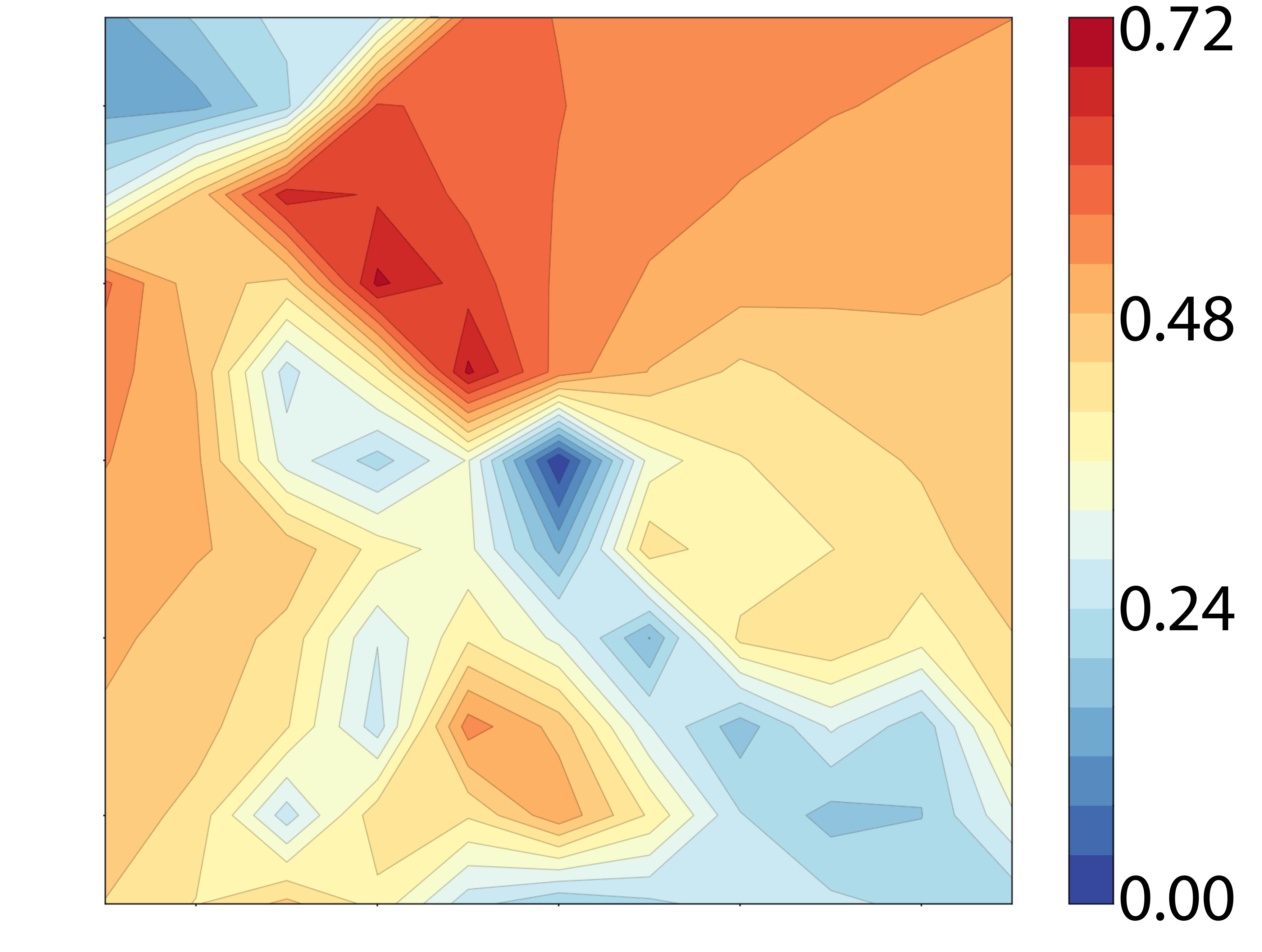}
    \caption{Seed 2 at 20k}
    \label{fig:multiberts_seed2_20k}
  \end{subfigure}\hfill
  \begin{subfigure}[t]{0.49\columnwidth}
    \centering
    \includegraphics[width=\columnwidth]{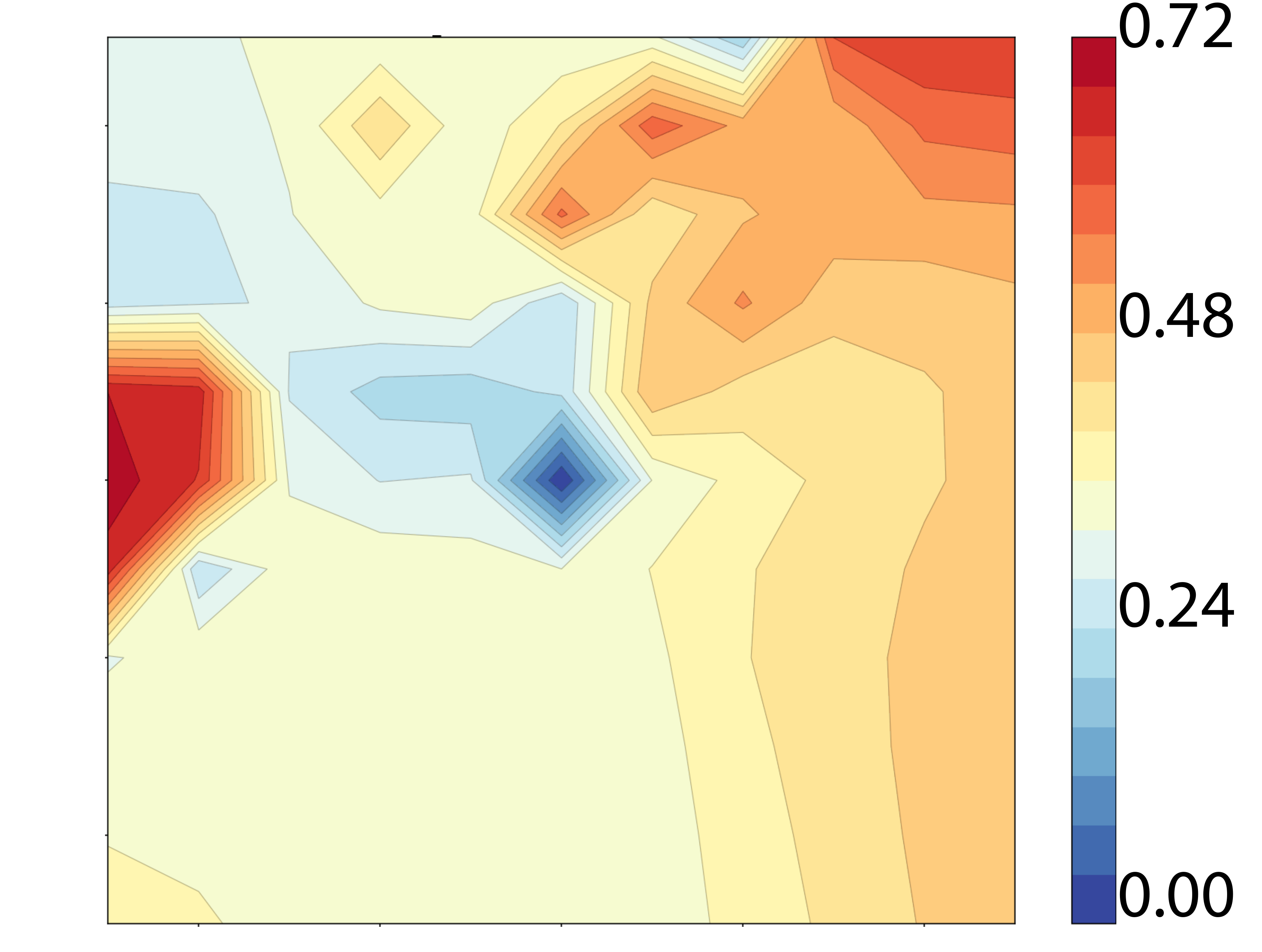}
    \caption{Seed 2 at 40k}
    \label{fig:multiberts_seed2_40k}
  \end{subfigure}

  \begin{subfigure}[t]{\columnwidth}
    \centering
    \includegraphics[width=0.8\columnwidth]{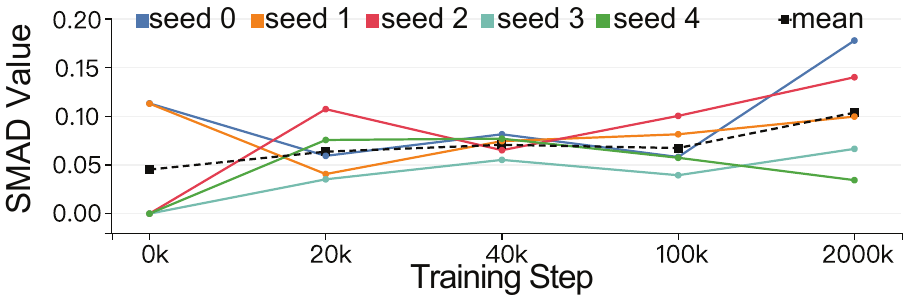}
    \caption{\textsc{SMAD} values across steps}
    \label{fig:multiberts_smad}
  \end{subfigure}


  \begin{subfigure}[t]{\columnwidth}
    \centering
    \includegraphics[width=0.8\columnwidth]{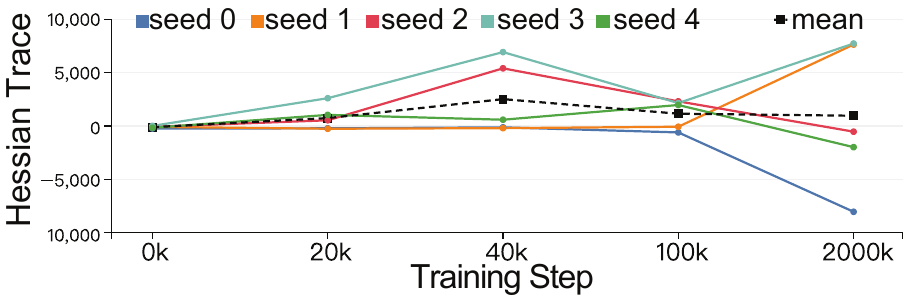}
    \caption{Hessian trace across steps}
    \label{fig:multiberts_trace}
  \end{subfigure}

  \caption{
    MultiBERT loss landscape analysis (Seed 2).
    \textbf{(a)-(b)} Contour plots of the 2D loss landscape show landscape smoothing from 20k to 40k steps.
    \textbf{(c)-(d)} Quantitative metrics diverge: \textsc{SMAD} decreases (consistent with smoothing), while the Hessian trace increases due to local curvature sharpening.
  }
  \label{fig:multiberts_analysis}
\end{figure}

This experiment highlights a critical distinction between local curvature metrics and more global topological complexity. To illustrate this divergence, Figure~\ref{fig:multiberts_analysis} (top) zooms in on a critical transition phase during the 2,000k step pre-training: the interval between 20k and 40k steps for Seed 2. 
Visually, the landscape undergoes a smoothing process, transitioning from a complex surface with irregularities to a more cohesive basin. \textsc{SMAD} accurately reflects this simplification, decreasing significantly as the topological noise resolves. 
However, as shown in Figure~\ref{fig:multiberts_analysis} (bottom), the trace of the Hessian actually \textit{increases} during this interval. 
This divergence highlights a crucial distinction: the Hessian trace strictly measures \textit{local curvature} (steepness), whereas \textsc{SMAD} quantifies \textit{global topological complexity} (roughness and connectivity). 
In this phase, the model settles into a steeper basin (increasing the Hessian trace) that is structurally simpler and less noisy (decreasing \textsc{SMAD}). 
This observation aligns with the taxonomy of loss landscapes proposed by~\citet{yang2021taxonomizing}, which posits that "sharp" minima can still possess simple global geometry. 
Unlike curvature metrics, \textsc{SMAD} explicitly measures this global structural simplification via the merge tree, distinguishing between landscapes that are simply "sharp" versus those that are topologically "chaotic."

\begin{table}[ht]
\centering
\caption{
    Multi-metric landscape analysis of MultiBERTs (Seed 2) during the critical pre-training phase. 
    To address the taxonomy proposed by \citet{yang2021taxonomizing}, we examine the landscape through distinct lenses: 
    (1) \emph{Dominant Curvature} ($\lambda_{max}$), indicating the sharpness of the steepest direction; 
    (2) \emph{Total Curvature} ($\htrace$), indicating average local sharpness; and 
    (3) \emph{Global Topology} (\textsc{SMAD}), indicating barrier height and basin connectivity.
    While spectral metrics ($\lambda_{max}$ and $\htrace$) indicate drastic sharpening, the topological metric (\textsc{SMAD}) indicates smoothing. 
    This confirms that the model enters a "Sharp-but-Connected" regime, a state that necessitates a topological metric like \textsc{SMAD} to be correctly identified.
}
\label{tab:multiberts_combination}
\begin{adjustbox}{max width=\columnwidth}
\begin{tabular}{l c c c c}
    \toprule
    & \multicolumn{2}{c}{\textbf{Local / Spectral Metrics}} & \textbf{Global / Topological Metric} & \\
    \cmidrule(lr){2-3} \cmidrule(lr){4-4}
    & \textbf{Max Eigenvalue} & \textbf{Hessian Trace} & \textbf{\textsc{SMAD}} & \\
    \textbf{Step} & $\lambda_{max}$ & $\htrace$ & Barriers \& Connectivity & \textbf{Landscape Taxonomy} \\
    \midrule
    20k & 5.53 & 533.63 & 0.107 & \textit{Rough \& Moderate} \\
    40k & 24.11 & 5412.79 & 0.065 & \textbf{\textit{Sharp \& Connected}} \\
    \midrule
    \textbf{Trend} & $\uparrow$ 4.3x (Sharper) & $\uparrow$ 10x (Sharper) & $\downarrow$ 40\% (Smoother) & \textbf{Validates Optimization} \\
    \bottomrule
\end{tabular}
\end{adjustbox}
\end{table}

\begin{figure*}[t] 
    \centering	
    \includegraphics[width=\textwidth]{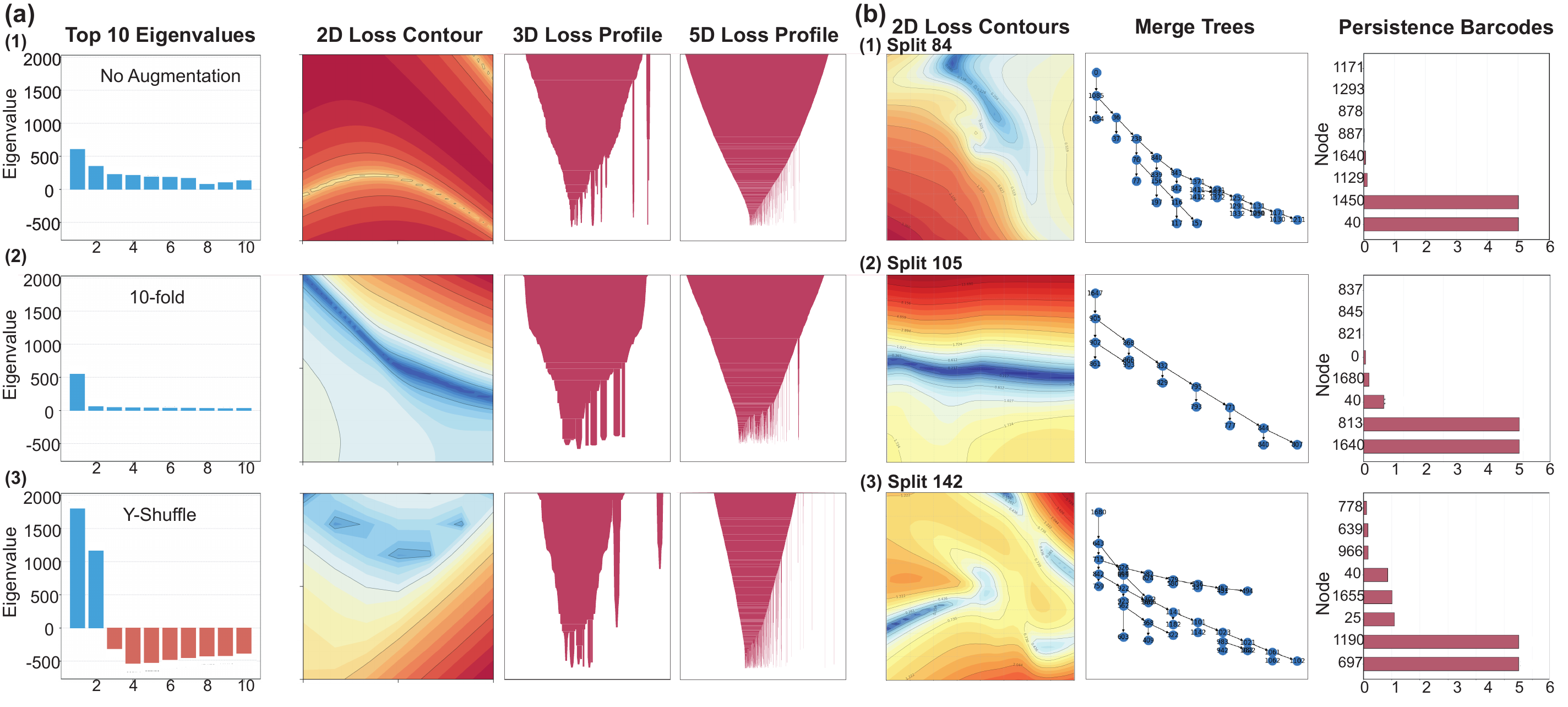}
    \caption{(a) $2\text{D}$ RMSE loss contours and 3D/5D landscape profiles derived from top Hessian eigenvectors, illustrating 3 distinct eigenvalue patterns: (1) similar top values, (2) sharp drop after the first eigenvalue, and (3) significant negative eigenvalues. (b) Merge trees and persistence barcodes, paired with their corresponding $2\text{D}$ RMSE loss contours on split sets of DimeNet++ train with 10-fold augmentation (seed $0$).
    } 
\label{fig:experiments_tda}
\end{figure*}

\subsection{Importance of Topological Data Analysis in Complex High-Dimensional Loss Landscapes}
\label{subsec:tda_highdim}

To illustrate the usefulness of the high-dimensional loss landscape visualization methods introduced in \Cref{sec:method}, we analyze the relationships among eigenvalue spectra, principal Hessian directions, and loss landscape features in several SciML models.
Specifically, we examine 3D GNNs trained on a real-world synthetic chemistry dataset to predict catalyst selectivity.~\citep{Hadler2025Olefin}. We examine several models differing in their data augmentation strategies, feature engineering choices, and predictive performance across two datasets. Each model was trained and validated on an ID literature-derived dataset, while OOD generalization was evaluated using an independent experimental dataset. Together, these models exemplify key challenges in practical SciML, including limited data availability, complex and highly nonconvex loss surfaces, and imperfect convergence.

The first two rows of \cref{fig:experiments_tda}(a) depict models with two distinct eigenvalue distributions: one smoothly decaying, and another dominated by a prominent leading eigenvalue. For Row 2, increasing the analysis dimension negligibly alters observations because the geometry is largely determined by the dominant eigenvector. However, for models with more evenly distributed eigenvalues (Row 1), the interpretation of the landscape changes strictly with dimensionality. 
For example, distinct basins of attraction (visualized as separate branches, representing the evolution of a connected component) that appear disconnected in a 3D projection are revealed to be part of a single connected valley in 5D. 
Crucially, this indicates that the barriers observed in 3D are artifacts of insufficient dimensionality, rather than intrinsic features of the loss surface. 
While any subspace projection is an approximation, the 5D analysis, by spanning more dominant Hessian directions, captures the landscape's global connectivity structure with significantly higher fidelity. 
By analyzing higher dimensions, we resolve projection-induced ambiguities, distinguishing between phantom barriers caused by aggressive dimensionality reduction and genuine obstacles to optimization. 

While eigenvalues are typically non-negative near minima in convex problems, negative eigenvalues frequently occur during SciML training, especially in data-limited regimes. These indicate saddle points or local maxima, representing directions of instability. 
However, a low-dimensional slice often excludes these specific directions, misrepresenting the nature of these critical points. 
For instance, a 2D projection might depict a region as a steep, insurmountable barrier (suggesting poor convergence), whereas the 5D analysis, by explicitly incorporating these directions of negative curvature, reveals it to be a saddle point connecting to a deeper minimum. 
By capturing this high-dimensional connectivity, \texttt{\PACKAGENAME} prevents the misinterpretation of landscape geometry, distinguishing between apparent optimization traps (artifacts of projection) and traversable saddle points (revealed by the principal subspace).

These observations indicate that the loss landscapes of 3D GNNs trained on data-limited chemical reaction prediction tasks are highly complex, in a domain interpretable way, exhibiting intricate basin structures, non-trivial connectivity, and significant variations in local curvature. 
This limitation underscores the importance of TDA techniques for probing higher-dimensional loss landscapes, as they can capture topological structure more faithfully than traditional analyses.

\paragraph{Quantifying Loss Landscape Geometry via TDA.} 
While loss landscape visualizations can provide intuitive observations, visual inspection alone often obscures fine-grained structural details. In this work, we complement these standard visualizations by leveraging topological descriptors, specifically merge trees and persistence barcodes, to characterize the landscape's multiscale geometry. Unlike scalar metrics that compress all information into a single number, these descriptors provide a structured summary of basin connectivity and feature persistence. This approach makes it possible to distinguish between meaningful geometric structures and topological noise, facilitating a more nuanced comparison across models. 

Contour plots (\cref{fig:experiments_tda}(b), first column) reveal intricate details, but analyzing basin depth and connectivity is challenging. In contrast, merge trees (\cref{fig:experiments_tda}(b), second column) more accurately depict connections between minima and saddle points. For instance, while contour plots (\cref{fig:experiments_tda}(b), first column) suggest Members 1 and 2 (Rows 1 and 2) are structurally similar because of their consistent smoothness, merge trees show that they are quantitatively more similar (shape, root/leaf distributions) to each other than to Member 3 (Row 3). This similarity, missed by contour plots, highlights the value of merge trees, which also distinguish Member 3 and the differences between Members 1 and 2.

However, merge trees alone do not directly quantify landscape smoothness, a property widely recognized as crucial for generalizability \citep{keskar2016large,yao2020pyhessian,YGSKM20_adahessian_TR_v1,foret2020sharpness}. 
To address this, we use persistence barcodes (\cref{fig:experiments_tda}(b), third column), an established TDA tool (\cref{apd:method_visualization}) that tracks the persistence of topological features (\cref{sec:background}). 
By definition, a bar's length corresponds to the persistence of the topological feature, representing the functional difference, or barrier height, required to exit a local minimum. 
Thus, an ideal smooth, single-basin landscape produces no short bars, as it contains no shallow local sub-optima to disrupt optimization. An abundance of bars can indicate noise caused by numerous shallow minima. These short-lived features can trap models and promote overfitting. The persistence barcodes (\cref{fig:experiments_tda}(b), third column) show that Member 3 is the noisiest, while Members 1 and 2 exhibit similar levels of noise. However, Member 1's landscape features are more detrimental to optimization because they contain an additional long bar, which represents a deep local minimum with a high energy barrier that can effectively trap the model. These details are not apparent in contour plots or merge trees. Thus, even in low-dimensional loss landscapes, TDA captures information more accurately than contour~plots.

\subsection{Assessing ID and OOD Generalization with SMAD}

Assessing a model’s ability to generalize beyond training data remains a central challenge in ML, especially in data-scarce domains such as chemistry and other areas of SciML. Traditional metrics such as RMSE or $R^2$ quantify predictive accuracy on known data but provide limited insight into a model’s ability to extrapolate beyond the training distribution. Here, we demonstrate how \textsc{SMAD} can serve as an indicator of a model's potential to generalize on OOD data.

To evaluate \textsc{SMAD} as an indicator of OOD generalization, we computed \textsc{SMAD} for 3D GNNs trained with varying levels of data augmentation and feature-augmentation choices from \citet{Hadler2025Olefin}, using loss landscapes derived from the literature ID test dataset. We then compared these values to losses on the experimental test dataset, which serves as an OOD benchmark relative to the literature training set, as shown in \cref{tab:smad_ID_ood}. Across these configurations, \textsc{SMAD} computed from literature-derived ID landscapes correlated strongly with OOD RMSE (Spearman’s $\rho = 0.928$, $p = 0.008$). Notably, \textsc{SMAD} correctly identifies the 10-fold augmented model (the variant with the lowest OOD RMSE) as having the lowest \textsc{SMAD} value, indicating that moderate data augmentation yields a well-regularized model capable of generalizing beyond its training distribution. Additionally, when adding the $\log_{10}(\textsc{SMAD})$ term to a baseline model using ID RMSE, the leave-one-out cross-validation (LOOCV) RMSE for predicting OOD RMSE was substantially reduced ($0.135 \rightarrow 0.020$; $\Delta = 0.115$, exact permutation test $p = 0.0056$). While the small sample size warrants caution, this result indicates that \textsc{SMAD} contributes predictive signal beyond ID RMSE and captures aspects of loss-landscape geometry that are directly relevant to OOD generalization.

These results provide evidence that \textsc{SMAD} could serve as a metric for OOD generalization by quantifying the smoothness of a model’s ID loss landscape, consistent with prior qualitative observations~\citep{iyer2021wideminimadensityhypothesisexploreexploit}, and thus offers actionable insight even without access to an explicit OOD test set. This is especially useful for SciML, where the difficulty of data collection can preclude assembling larger or truly OOD external datasets.

\begin{table}[ht]
    \centering
    \small
    \caption{\textsc{SMAD} values for 3D GNNs trained with different levels of data augmentation and feature engineering to predict catalyst selectivity from \citet{Hadler2025Olefin}, along with their corresponding test losses on an ID literature dataset and an OOD experimental dataset.}
    \label{tab:smad_ID_ood}
    \begin{tabular}{lccccc}
        \toprule
        Model & \makecell{ID \\ RMSE} & \makecell{ID \\ $R^2$} & \makecell{OOD \\ RMSE} & \makecell{OOD \\ $R^2$} & \textsc{SMAD} \\
        \midrule
        No Aug.   & 0.80 & 0.61 & 0.57 & 0.65 & $6.8 \times 10^{-4}$ \\
        \textbf{{10-Fold}}   & \textbf{{0.64}} & \textbf{{0.75}} & \textbf{{0.50}} & \textbf{{0.74}} & \textbf{{\(\bm{1.7 \times 10^{-4}}\)}} \\
        20-Fold   & 0.67 & 0.72 & 0.59 & 0.62 & $2.2 \times 10^{-3}$ \\
        100-Fold  & 0.67 & 0.72 & 0.56 & 0.66 & $7.2 \times 10^{-4}$ \\
        Charge    & 0.64 & 0.75 & 0.50 & 0.73 & $2.0 \times 10^{-4}$ \\
        xTB Topo. & 0.65 & 0.74 & 0.55 & 0.67 & $4.7 \times 10^{-4}$ \\
        \bottomrule
    \end{tabular}
\end{table}

\section{Discussion and Conclusion}
\label{sec:discussion}

This work introduced \texttt{\PACKAGENAME}, a unified framework for constructing, visualizing, and quantifying loss landscapes.
We propose \textsc{SMAD}, a novel topological metric designed to measure landscape smoothness and connectivity. 
By integrating Hessian-based subspace construction with TDA, \texttt{\PACKAGENAME} reveals geometric structures, such as basin hierarchy and connectivity, that are undetectable in traditional low-dimensional projections.

Through cross-architecture studies spanning CNNs, Transformers, and GNNs, we demonstrated that \textsc{SMAD} captures interpretable differences in optimization geometry that local curvature metrics miss. Our results underscore the usefulness of more global topological metrics, as they distinguish between locally sharp yet globally well-connected basins and truly pathological optimization landscapes.

Finally, we demonstrated the practical value of this approach in SciML, where data scarcity precludes extensive validation sets. In real-world chemical reaction prediction tasks, predicting catalyst selectivity, \textsc{SMAD} successfully identified OOD generalization potential solely from ID landscape geometry. By linking smoother, more connected landscapes with improved extrapolative performance, \textsc{SMAD} provides a promising proxy for assessing model robustness. Together, these results establish \texttt{\PACKAGENAME} as a practical toolkit for diagnosing model behavior in ML and SciML, guiding architecture design, and bridging the gap between local curvature analysis and global topological structure.

\section{Limitations and Future Work}
\label{sec:limitations_future_work}

While our framework offers novel insights into loss landscape analysis, it also has limitations that motivate clear directions for future research.

\paragraph{Interpretability and Metric Robustness.} Topologically derived metrics such as \textsc{SMAD} are most informative when interpreted jointly with complementary measures. 
Accordingly, integrated multi-metric analyses are essential for robustness, especially in noisy optimization settings.

\paragraph{Validation.} Although we demonstrate applicability across CNNs, Transformers, and GNNs, further validation is needed to establish generalizability across additional domains, architectures, and data regimes, particularly in large-scale training environments.

\paragraph{Computational Considerations and Scalability.} Our method involves a trade-off between analysis dimensionality and sampling resolution, driven by three main bottlenecks: (i) \emph{Hessian computation}, which requires $O(d^3)$ time and $O(d^2)$ memory to extract eigenvectors at each grid point; (ii) \emph{grid sampling}, where the number of sampled points grows exponentially with dimensionality as $k^d$; and (iii) \emph{topological analysis}, whose TDA runtime scales with the total number of samples, which we address for $d>3$ using sparse filtrations and parallelization. These constraints introduce an inherent tradeoff: higher-dimensional analyses provide richer geometric insight but require sparser sampling, which can miss fine-grained topological structure.

\begin{table}[!ht]
\centering
\caption{Approximate run times for loss-landscape analysis tasks using a single GPU (NVIDIA A100 40GB).}
\begin{tabular}{l r}
\toprule
Task & Runtime (s) \\
\midrule
Calculate Top 3 Hessian Eigenvectors & 109.82 \\
Generate 2D Loss Landscape           & 72.52  \\
Generate 3D Loss Landscape           & 926.58 \\
\bottomrule
\end{tabular}
\label{tab:runtime-breakdown}
\end{table}

\paragraph{Future Directions.} These limitations motivate several extensions. We are expanding \texttt{\PACKAGENAME} to other data-scarce SciML applications and exploring strategies for improved scalability, such as nonlinear mode connectivity and graph-based topological compression. In parallel, we aim to develop a unified multi-metric framework that integrates \textsc{SMAD} with auxiliary measures to improve interpretability and statistical robustness. Addressing these challenges will advance scalable, high-dimensional, and interpretable landscape analysis for complex deep learning systems.

\section{Code Availability}
\texttt{Landscaper} is open-source and available at \url{https://github.com/Vis4SciML/Landscaper}. 
A stable release is available on PyPI at \url{https://pypi.org/project/Landscaper/}, and documentation is hosted at \url{https://vis4sciml.github.io/Landscaper/}.

\section{Impact Statement}
\label{sec:impact}

This paper presents work whose goal is to advance the field of Explainable and Trustworthy Machine Learning. By introducing \texttt{\PACKAGENAME}, an open-source framework for arbitrary-dimensional loss landscape analysis, we aim to enhance the transparency and interpretability of complex ``black-box'' neural networks. 

Our topological approach, particularly the \textsc{SMAD} metric, allows practitioners to diagnose optimization geometry and assess OOD generalization potential using only ID data. This contribution has significant positive societal impacts, particularly in high-stakes Scientific Machine Learning (SciML) domains (e.g., chemical property prediction and material science), where data is often scarce, and model reliability is critical for safety and scientific discovery. 

By enabling researchers to distinguish between locally sharp but globally well-connected basins and truly pathological landscapes, our tool fosters the development of more robust, accountable, and fair AI systems. As an analytical and diagnostic toolkit, we do not foresee any immediate negative societal consequences or ethical risks resulting directly from our theoretical or methodological contributions.

\section{Acknowledgments}
The authors would like to thank Arnur Nigmetov (Lawrence Berkeley National Laboratory) for his valuable guidance and suggestions on the work related to Topological Data Analysis (TDA). This work was supported by the Laboratory Directed Research and Development (LDRD) Program of Lawrence Berkeley National Laboratory and by the U.S. Department of Energy, Office of Science, Advanced Scientific Computing Research (ASCR) program under Contract No. DE-AC02-05CH11231 to Lawrence Berkeley National Laboratory and Award No. DE-SC0023328 to Arizona State University (“Visualizing High-dimensional Functions in Scientific Machine Learning”). 
We would also like to acknowledge the DOE Competitive Portfolios grant and the DOE SciGPT grant.
This research used resources at the National Energy Research Scientific Computing Center (NERSC), a U.S. Department of Energy, Office of Science, User Facility under NERSC Award Number ASCR-ERCAP0026937. We thank the NIH for funding this work at UC Berkeley (1R35GM130387). N.H. was supported by the National Science Foundation Graduate Research Fellowship Program under Grant No. (DGE 2146752). Any opinions, findings, and conclusions or recommendations expressed in this material are those of the author(s) and do not necessarily reflect the views of the National Science Foundation.

\bibliographystyle{plainnat}
\bibliography{example_paper}

\clearpage
\appendix
\crefalias{section}{appendix}
\crefalias{subsection}{appendix}
\crefalias{subsubsection}{appendix}
\onecolumn
\section{Methods}
\label{apd:method}

This appendix provides the technical specifications and mathematical foundations for the \texttt{\PACKAGENAME} framework. While the complete workflow integrates three core components: subspace construction, topological visualization, and the \textsc{SMAD} quantification metric. This appendix focuses on algorithmic implementations of the first two. Specifically, we present the construction of high-dimensional landscape samples via Hessian eigenvectors (\Cref{apd:method_construction}) and the generation of topological visualizations for interpretability (\Cref{apd:method_visualization}).

\subsection{Construction: High-Dimensional Landscape Sampling}
\label{apd:method_construction}

Here, we detail the process of projecting the loss landscape into an $n$-dimensional subspace spanned by the top $n$ Hessian eigenvectors, which serves as the basis for our subsequent topological analysis.

\texttt{\PACKAGENAME} extend from two dimensions to $n$ dimensions by calculating the top $n$ Hessian eigenvectors using \texttt{PyHessian} \citep{yao2020pyhessian} and sampling along the subspace they span. The idea is that by leveraging the eigenvectors associated with the top $n$ largest eigenvalues, we can visualize the most significant local loss fluctuations for a given model. This approach is supported by findings, such as those in~\citep{gur2018gradient}, which show that the dynamics of training are largely confined to a low-dimensional subspace spanned by the leading Hessian eigenvectors, suggesting that these directions capture critical curvature information. Similarly,~\citet{papyan2020traces} motivates characterizing the Hessian using only the top few hundred directions, rather than the full parameter space, highlighting that the number of significant eigenvalues (``spikes'') may be proportional to the number of classes. This justifies focusing on a small number of top eigenvectors and avoids the computational cost of analyzing the full Hessian spectrum. Formally, we extend the traditional definition of $2\text{D}$ loss landscapes to perturb model parameters along $n$ directions and evaluate the loss $\mathcal{L}$ as follows: 
\begin{equation}
  \label{eq:projection_highdim}
  f(\alpha_1 ... \alpha_n) = \mathcal{L} (\theta + \Sigma_{i=1}^{n}\alpha_i\delta_i)  ,
\end{equation}
where $\alpha_1, ..., \alpha_n$ are the coordinates within an $n$-dimensional subspace, $\delta_i$ denotes the $i$-th eigenvector direction of the Hessian, and $\theta$ represents the original model parameters. Each coordinate $(\alpha_1, ..., \alpha_n)$ corresponds to a point in parameter space with an associated loss value, and, together, these values define an $n$-dimensional loss landscape.

\texttt{\PACKAGENAME} constructs a landscape with a coordinate meshgrid; each coordinate $\alpha_i$ is sampled from a set of defined values centered at the original model's parameters. 
In the experiments presented in this work, we employ a uniform range $[-r, r]$ and fixed resolution across all dimensions to ensure consistent comparisons. 
However, the framework is designed to be fully flexible: it supports anisotropic grids where the range $[-r_i, r_i]$ or the sampling density can be individually scaled based on the curvature (e.g., proportional to the inverse of the corresponding eigenvalue $\lambda_i^{-1/2}$). 
The final sampling grid is the Cartesian product of these coordinate sets, allowing users to tailor the exploration granularity to the specific geometry of the Hessian eigenvectors.

Although grid sampling scales exponentially with dimensionality, our empirical results operate in a moderately high-dimensional regime (typically $n \leq 5$), where it remains effective for revealing key topological features. This framework is theoretically agnostic to the sampling method, allowing it to be extended to higher dimensions using alternative sampling strategies, e.g., random or adaptive sampling. 
Our primary aim is to highlight the importance of capturing high-dimensional structure in loss landscapes, not to compare sampling strategies. 
Tools like TDA enable principled analysis beyond traditional low-dimensional projections, offering deeper guidance into the optimization landscape of deep learning models.

\subsection{Visualization: Interpreting the Loss Landscape}
\label{apd:method_visualization}

Here, we describe the specific visualization techniques implemented in \texttt{\PACKAGENAME}, loss landscape contours, landscape merge trees, landscape profiles, and persistence barcodes, and explain how they interpret the geometric features constructed in the previous subsection.

\begin{figure*}[t]
    \centering	
    \includegraphics[width=0.8\linewidth]{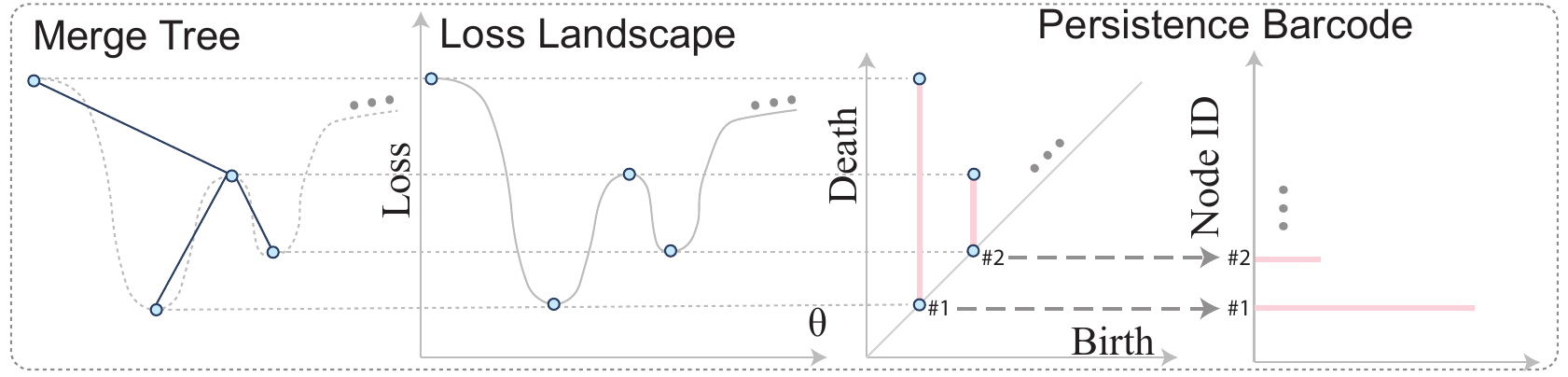}
    \caption{
    Two standard TDA visualization methods are employed in \texttt{\PACKAGENAME}: the merge tree and the persistence barcode. Critical points are extracted from the loss landscape and are organized into a merge tree with their persistence values; minima are connected with their corresponding saddle-points. The persistence barcode can be seen as a 1D projection of the merge tree, which displays the life-spans of the connected components formed by saddle-minima pairs, starting from their appearance (birth) to when they merge (death).
    }
\label{fig:mergetree_persistencebarcode}
\end{figure*}

\begin{figure*}[t]
    \centering	
    \includegraphics[width=\linewidth]{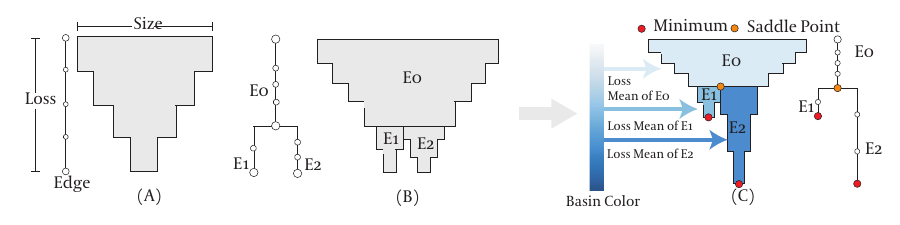}
    \caption{
    Profile Visualization. Representing a merge tree as a topological profile. (A) shows a single basin corresponding to a merge tree with a single branch, and (B) shows multiple basins corresponding to multiple branches. (C) demonstrates the color scheme.
    }
\label{fig:topological_landscape_profile_construction}
\end{figure*}

\paragraph{Profile Visualization.} Scalar quantities often provide only a limited view of a model’s behavior. Visualization serves as a crucial component for analysis, offering a more holistic and interpretable perspective on the optimization dynamics and generalization properties of a model. By revealing geometric and topological features of high-dimensional loss landscapes, visualizations can uncover patterns and behaviors that may be obscured by aggregate statistics \citep{tufte1983visual, healy2024data}. This makes visualizations particularly valuable for diagnosing training instability, comparing architectures, or understanding the effects of regularization and optimization strategies. 

To achieve this goal, \texttt{\PACKAGENAME} provides a visualization method using the landscape profile approach proposed by~\citep{geniesse2024visualizing} based on the landscape's merge tree (\cref{fig:topological_landscape_profile_construction}). 
To make high-dimensional structures interpretable, the profile visualization summarizes key features, such as flatness, sharpness, and curvature anisotropy. 
Specifically, this method projects the merge tree's topological skeleton onto a 2D plane while employing statistical volume estimation to determine the width of each basin, thereby visually encoding the landscape's geometry.

\paragraph{Persistence Barcodes.} Converting loss landscapes into merge trees (\cref{fig:mergetree_persistencebarcode}; left) or landscape profiles (\cref{fig:topological_landscape_profile_construction}) is effective for capturing minima and saddle points in higher-dimensional landscapes. However, in local regions, the smoothness of the transitions, particularly from saddles to minima, can also reflect important characteristics of the model's behavior. 
To quantify this, we employ \textit{Persistence Barcodes} (\cref{fig:mergetree_persistencebarcode}; right), which track the ``lifespan'' of topological features. In this representation, short bars correspond to shallow, noisy fluctuations (roughness), while long bars indicate deep, robust basins. This distinction allows barcodes to provide a direct, quantitative measure of local landscape smoothness, resolving the fine-grained texture that merge trees may abstract away.

To address this limitation, we propose using persistence barcodes (shown in \cref{fig:mergetree_persistencebarcode}; right) as a complementary tool to the merge tree, which allows us to better characterize the local smoothness of the landscape, especially the presence of small branches between critical points. In this context, a landscape with many small branches indicates a reduced smoothness and more local variability. In contrast, the absence of such branches suggests overly sharp or abrupt transitions between critical points, which may also reflect a lack of local smoothness. By integrating persistence barcodes into our analysis, we capture a more complex view of the geometry of the local landscape that merge trees alone may overlook.

\section{Supplemental Figures}
\label{apd:supplemental_figures}

\subsection{CNN Loss Landscapes for Underfit, Well-Fit, and Overfit Regimes}
\label{apd:cnn_landscapes}

\begin{figure}[ht]
    \centering
    \begin{subfigure}[b]{0.49\linewidth}
        \centering
        \includegraphics[width=\linewidth]{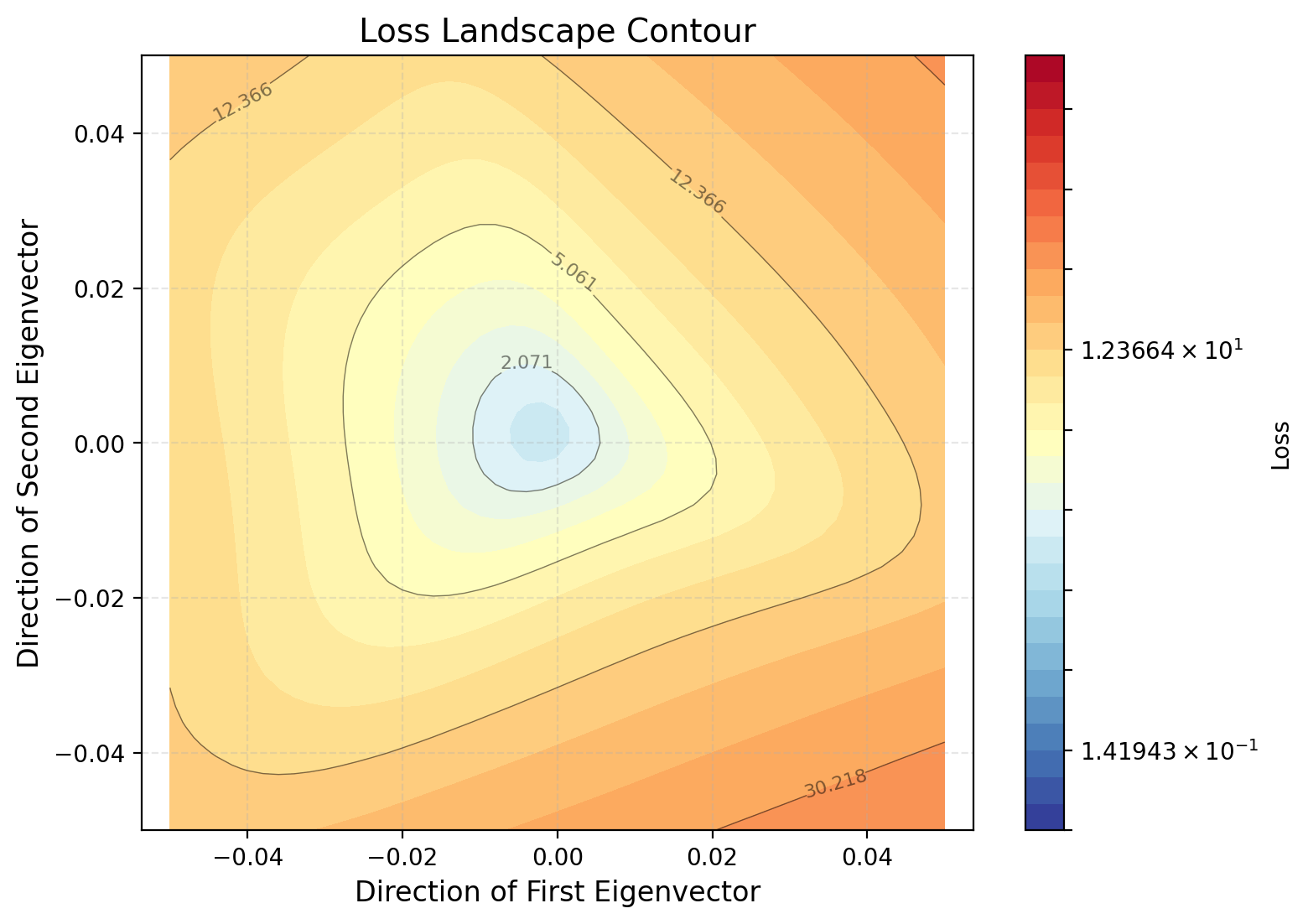}
        \caption{2D plot of the \textit{underfit} loss landscape.}
        \label{fig:underfit_2d}
    \end{subfigure}
    \hfill
    \begin{subfigure}[b]{0.49\linewidth}
        \centering
        \includegraphics[width=\linewidth]{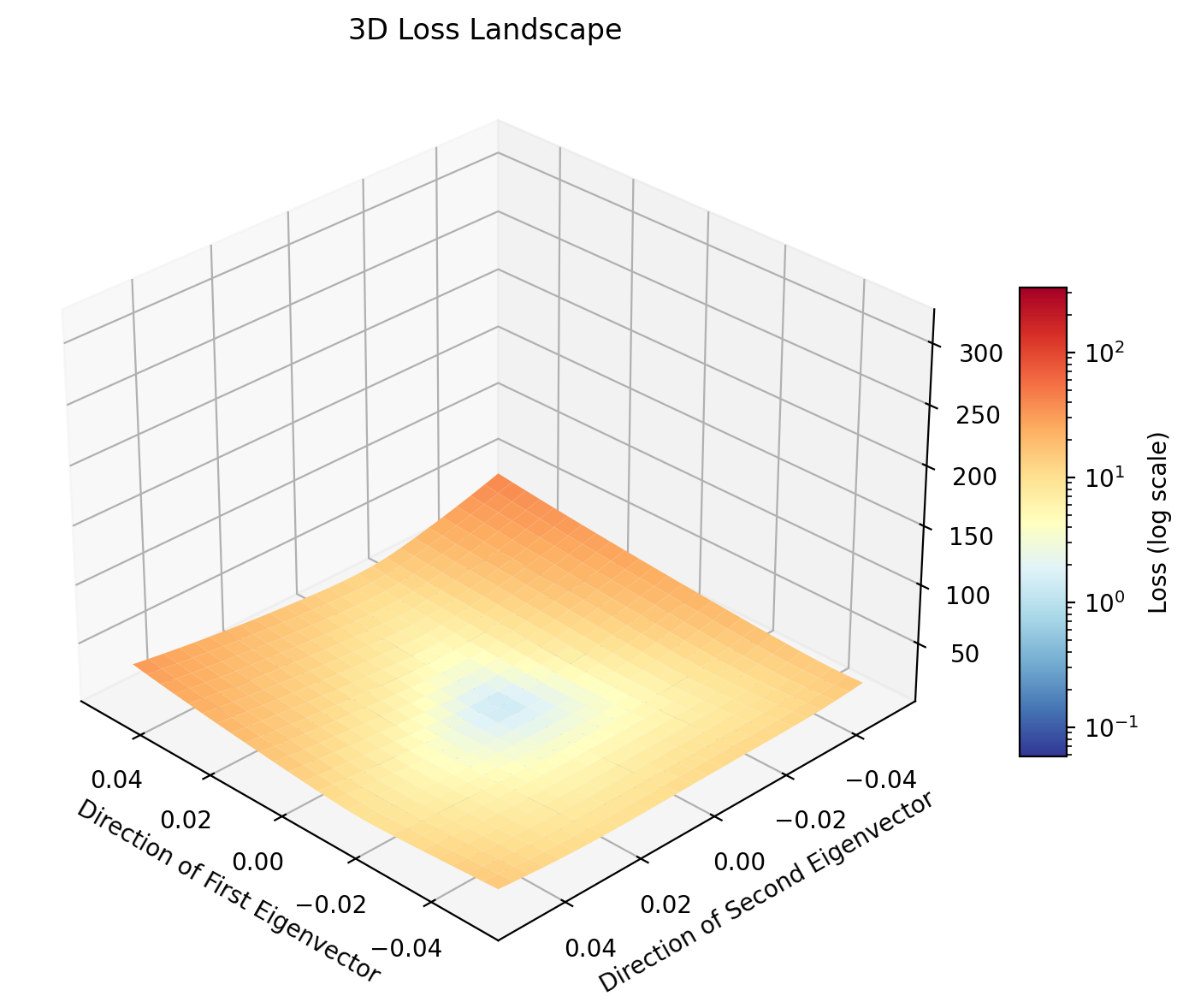}
        \caption{3D plot of the \textit{underfit} loss landscape.}
        \label{fig:underfit_3d}
    \end{subfigure}
    \caption{Loss landscapes of a simple CNN in an \textit{underfit} training regime trained on CIFAR-10. The landscape was computed using the top two Hessian eigenvector directions with a distance of 0.05 over 51 steps.}
    \label{fig:underfit_2d_3d}
\end{figure}

\begin{figure}[ht]
    \centering
    \begin{subfigure}[b]{0.49\linewidth}
        \centering
        \includegraphics[width=\linewidth]{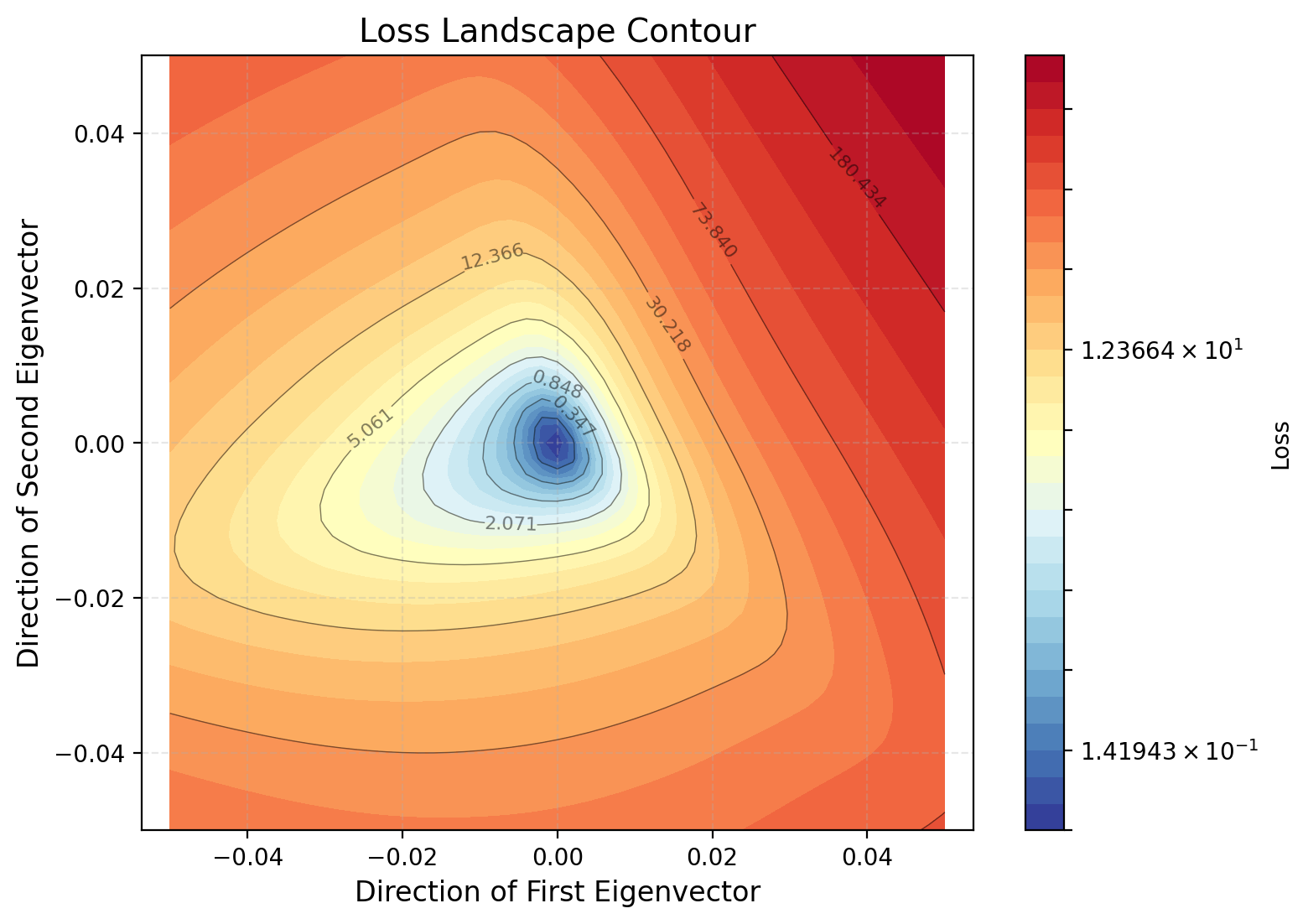}
        \caption{2D plot of the \textit{well-fit} loss landscape.}
        \label{fig:wellfit_2d}
    \end{subfigure}
    \hfill
    \begin{subfigure}[b]{0.49\linewidth}
        \centering
        \includegraphics[width=\linewidth]{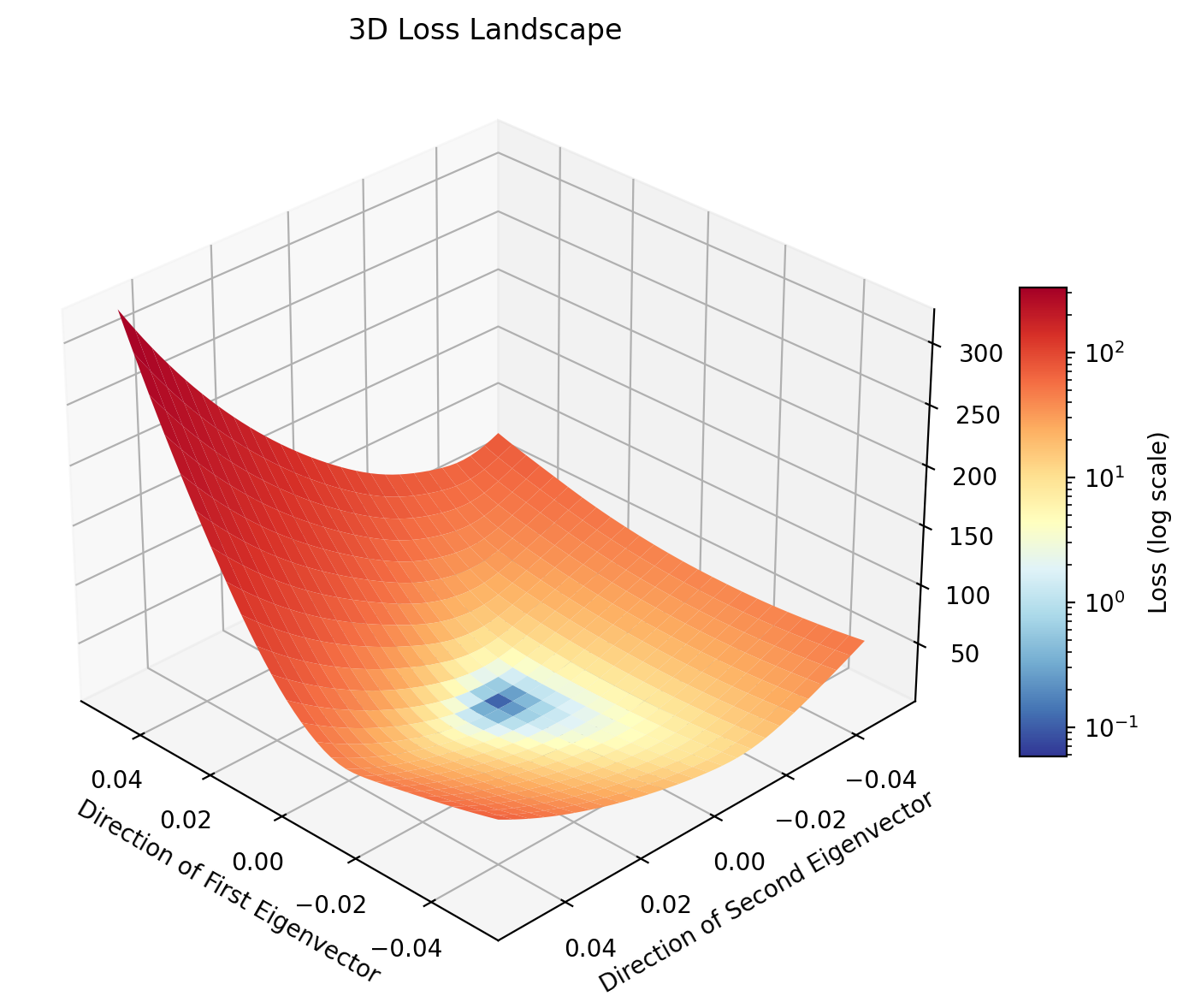}
        \caption{3D plot of the \textit{well-fit} loss landscape.}
        \label{fig:wellfit_3d}
    \end{subfigure}
    \caption{Loss landscapes of a simple CNN in an \textit{well-fit} training regime trained on CIFAR-10. The landscape was computed using the top two Hessian eigenvector directions with a distance of 0.05 over 51 steps.}
    \label{fig:wellfit_2d_3d}
\end{figure}

\begin{figure}[ht]
    \centering
    \begin{subfigure}[b]{0.49\linewidth}
        \centering
        \includegraphics[width=\linewidth]{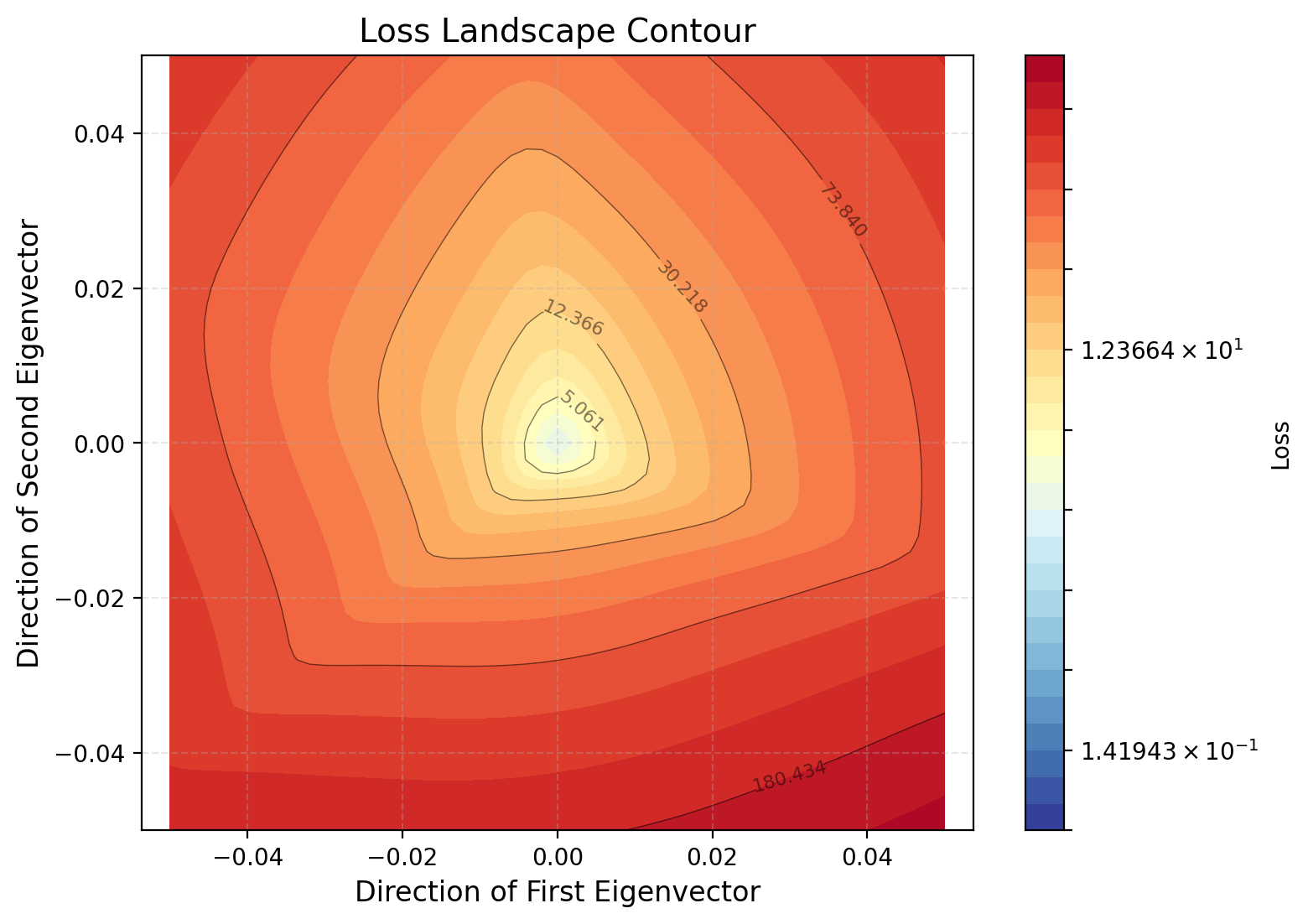}
        \caption{2D plot of the \textit{overfit} loss landscape.}
        \label{fig:overfit_2d}
    \end{subfigure}
    \hfill
    \begin{subfigure}[b]{0.49\linewidth}
        \centering
        \includegraphics[width=\linewidth]{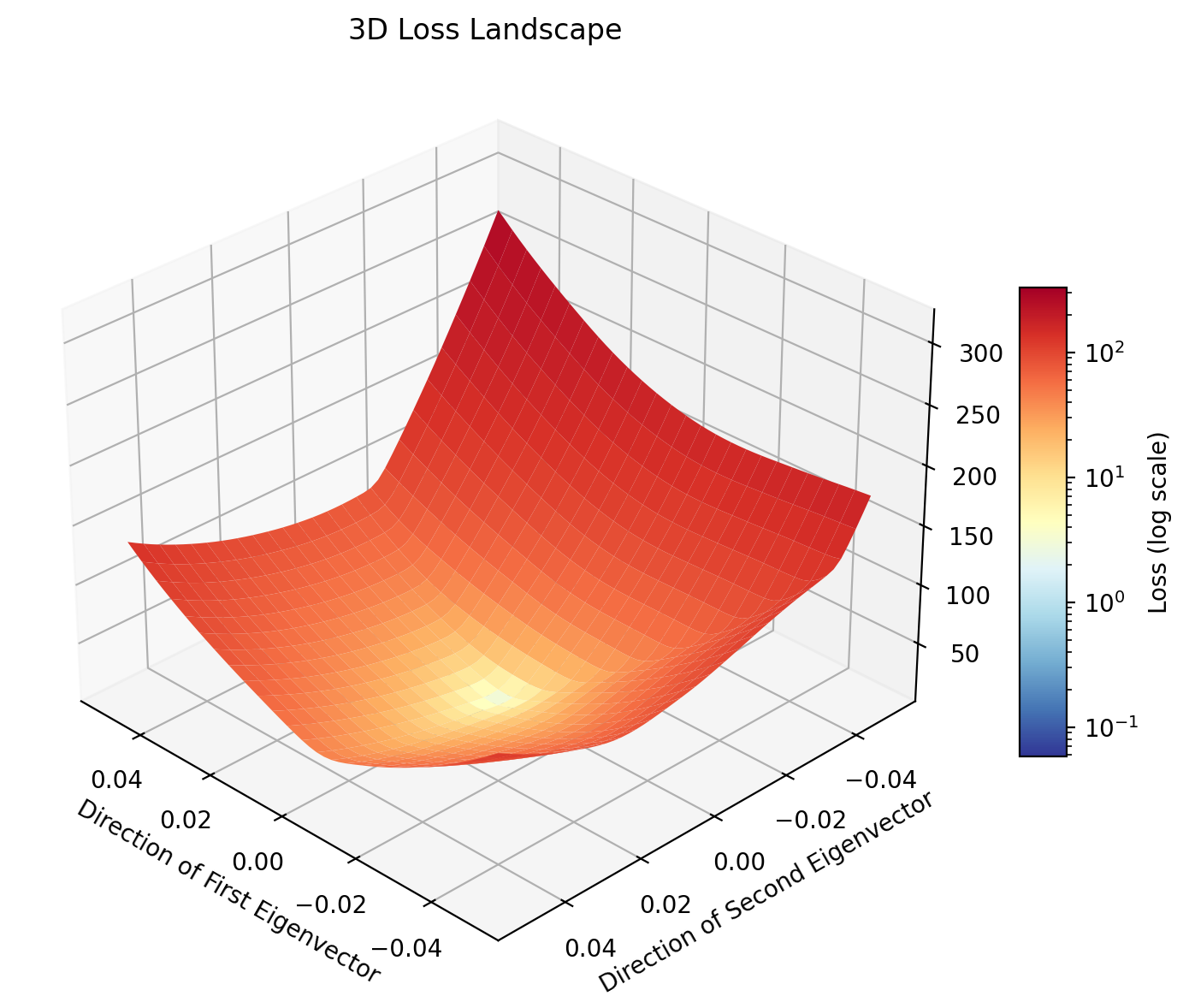}
        \caption{3D plot of the \textit{overfit} loss landscape.}
        \label{fig:overfit_3d}
    \end{subfigure}
    \caption{Loss landscapes of a simple CNN in an \textit{overfit} training regime trained on CIFAR-10. The landscape was computed using the top two Hessian eigenvector directions with a distance of 0.05 over 51 steps.}
    \label{fig:overfit_2d_3d}
\end{figure}

\end{document}